# Unsupervised Image-generation Enhanced Adaptation for Object Detection in Thermal Images

*Peng Liu[1], Fuyu Li[2], Wanyi Li[2]\*, Member, IEEE*
1. China National Institute of Standardization,
2. Institute of Automation, Chinese Academy of Sciences, {fuyu.li, wanyi.li}@ia.ac.cn
\* The corresponding author

## ABSTRACT

Object detection in thermal images is an important computer vision task and has many applications such as unmanned vehicles, robotics, surveillance and night vision. Deep learning based detectors have achieved major progress, which usually need large amount of labelled training data. However, labelled data for object detection in thermal images is scarce and expensive to collect. How to take advantage of the large number labelled visible images and adapt them into thermal image domain, is expected to solve. This paper proposes an unsupervised image-generation enhanced adaptation method for object detection in thermal images. To reduce the gap between visible domain and thermal domain, the proposed method manages to generate simulated fake thermal images that are similar to the target images, and preserves the annotation information of the visible source domain. The image generation includes a CycleGAN based image-to-image translation and an intensity inversion transformation. Generated fake thermal images are used as renewed source domain. And then the off-the-shelf Domain Adaptive Faster RCNN is utilized to reduce the gap between generated intermediate domain and the thermal target domain. Experiments demonstrate the effectiveness and superiority of the proposed method.

## 1. INTRODUCTION

Thermal cameras capture passively the infrared radiation emitted by all objects with a temperature above absolute zero [1]. Vision systems using thermal cameras can eliminate the illumination problems of normal greyscale and RGB cameras. Object detection in thermal images is a very important computer vision task, and has many applications including unmanned vehicles, robotics, surveillance, night vision, industrial, and military etc.

Deep learning based detectors, such as Faster RCNN [2], SSD [3], YOLO [4], have achieved major progress in visible domain, which usually need large amount of labelled training data. However, labelled thermal images for training object detectors are scarce and expensive to collect, while there are large amount of labelled visible images. Thus, it is expected to make use of these annotated visible images and adapt them into thermal image domain for object detection. This problem is referred as domain adaptive object detection from visible to thermal.

The research on object detection in thermal images under domain adaptation context is not as developed as that with color, including only several methods. [5] propose to transforms the thermal IR data as close as possible to the RGB domain via basic image processing operations, and fine-tune the pre-trained CNN based detector on preprocessed data. [6] presented an approach to pedestrian detection in thermal infrared images with limited annotations. The authors tackle the domain shift between thermal and color images by learning a pair of image transformers to convert images between the two modalities, jointly with a pedestrian detector. For general domain adaptive object detection, [7] is the first work to deal with the domain adaptation problem for object detection. The authors conducted adversarial training on features and designed three adaptation components to deal with domain shift, i.e., image-level adaptation, instance-level adaptation and consistency check.

Comparing to the above mentioned works, to our best knowledge, this paper is the first work to deal with unsupervised adaptive object detection from visible to thermal domain. The contributions of this work mainly consist of the following three aspects:

(1) We propose an unsupervised image-generation enhanced adaptation method for object detection in thermal images, in which, an image generation module and a re-adaptation module are included.

(2) To reduce the gap between visible domain and thermal domain, an image generation process is designed. The image generation process consists of a CycleGAN based image-to-image translation and an intensity inversion transformation.

(3) We conduct extensive experiments to compare the proposed methods with other methods, where it yields notable performance gains.

## 2. PROPOSED METHOD

In this section, we present details of our proposed unsupervised image-generation enhanced domain adaptive thermal object detector. Figure 1 shows the overview framework. It consists of two modules, image generation and re-adaptation. The image generation module generates simulated fake thermal images by a CycleGAN image translation process and an intensity inversion transformation. The re-adaptation module firstly takes the generated fake thermal images as renewed source domain, the real thermal as target domain. And then conducts an off-the-shelf Domain Adaptive Faster RCNN for object detection. Trained detector can be applied to the thermal target domain. More details are provided in the following subsections.

**Image generation**: generating fake thermal images

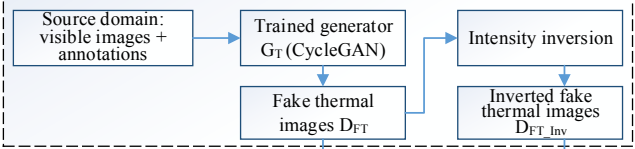

**Re-adaptation**: adapt from fake thermal domain to real thermal domain

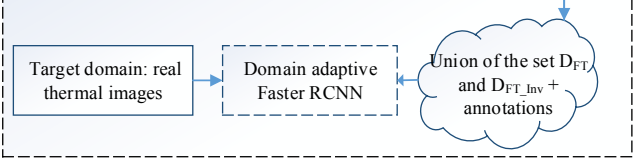

**Figure 1.** Overview framework of our proposed unsupervised image-generation enhanced adaptive thermal object detector.

### 2.1 Image generation

To reduce the gap between the visible source domain and the thermal target domain, we design an image generation module to generate simulated images that are similar to target images. The module consists two steps, a CycleGAN [8] step for translate visible image to thermal style, and an intensity inversion step to diversify the appearance of generated fake thermal images.

*2.1.1. Image translation via CycleGAN* [8]

CycleGAN is an unpaired image-to-image translation method. In this paper, the goal of CycleGAN [8] is to learn a mapping $G_T: V \rightarrow T$ such that the distribution of images from $G_T(V)$ is indistinguishable from the distribution $T$ using an adversarial loss. Because this mapping is highly under-constrained, $G_T$ is coupled with an inverse mapping $G_V: T \rightarrow V$ and introduce a cycle consistency loss to enforce $G_V(G_T(V)) \approx V$ (and vice versa). $V$ represents the color visible domain and $T$ represents the thermal domain. The objective of CycleGAN to minimize is shown as Eq.(1).

$$\mathcal{L}(G_T, G_V, D_V, D_T)$$
$$= \mathcal{L}_{GAN}(G_T, D_T, V, T)$$
$$+ \mathcal{L}_{GAN}(G_V, D_V, T, V) + \lambda \mathcal{L}_{cyc}(G_T, G_V)$$
(1)

In Eq.(1), $\mathcal{L}_{GAN}(G_T, D_T, V, T)$ and $\mathcal{L}_{GAN}(G_V, D_V, T, V)$ is the adversarial losses of mapping function $G_T$ and $G_V$ respectively, $\mathcal{L}_{cyc}(G_T, G_V)$ is the cycle consistency loss. $\lambda$ denotes the relative importance of the adversarial losses and cycle consistency loss. The optimization problem to solve is:

$$G_T^*, G_V^* = arg \min_{G_T, G_V} \max_{D_V, D_T} \mathcal{L}(G_T, G_V, D_V, D_T)$$
(2)

Translated fake thermal images for demonstration are shown in Figure 2. Images of the left column are from color visible domain, of the middle column are generated fake thermal images, and of the right column are real ground truth thermal images.

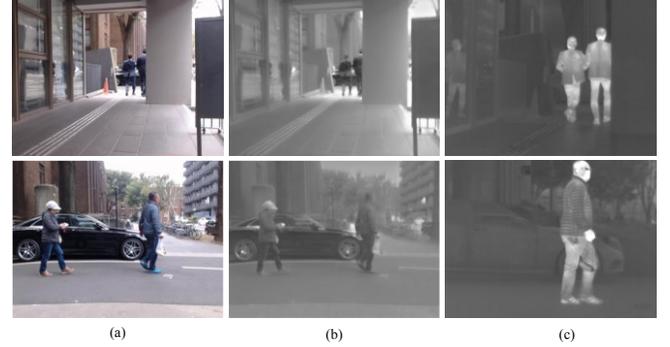

(a)         (b)         (c)

**Figure 2.** Generated fake thermal images for example. (a) Color visible images; (b) Generated fake thermal images; and (c) Real ground truth thermal images. The color images from top to down are 000492.png and 000505.png in RGB folder of Multi-spectral Object Detection dataset [9].

*2.1.2. Intensity inversion*

The generated fake thermal images and the real ground truth thermal images are compared in Figure 2 (b) and Figure 2 (c). It is likely that the generated fake thermal images are with the contents of the color visible domain images and with the style of the thermal domain images. However, the intensity of specific target object region is opposite, such as person region. From Figure 2 (b) and Figure 2 (c), it is shown that the intensity of person region in fake images is low while that of real thermal images is high. We argue that if we train detectors using only images similar to Figure 2 (b), the detector will miss the objects with inverse intensity. This argument is shown in our experiments, details can be found in the ablation study, i.e., Section 3.3.

Based on the above analysis, we propose to augment the generated fake thermal images by an intensity inversion transformation. The augmentation is expected to diversify the appearance of labelled training data and improve the performance of the object detector. The proposed intensity inversion transformation is defined as Eq. (3).

$$f_{inv}: T_{inv} = 255 - T$$
(3)

In Eq. (3), the invert function $f_{inv}$ corresponds to the intensity inversion transformation, $T$ denotes the fake thermal image to invert which is an eight bit image, and $T_{inv}$ denotes the inverted image.

Examples of intensity inversion transformation are shown in Figure 3. The appearance of object region in inverted images becomes similar to that of real thermal images.

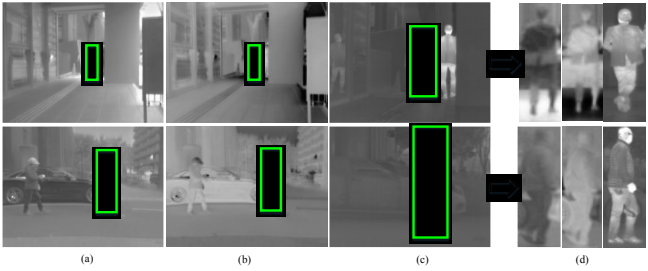

**Figure 3.** Illustration of intensity inversion transformation. (a) Generated fake thermal images; (b) Inverted fake thermal images; (c) Real ground truth thermal images, and (d) Cropped object regions. The images from top to down are 000492.png and 000505.png of Multi-spectral Object Detection dataset [9].

**2.2 Re-adaptation**
After done the image generation module, we take the union of generated fake thermal images and inverted fake thermal images as renewed source domain, which is defined as

$$S_{renewed}:\{D_{S_{renewed}}, B_{D_V} | D_{S_{renewed}} = D_{FT} \cup D_{FT_{Inv}}\} \quad (4)$$

where $S_{renewed}$ denotes the renewed source domain, which consists of the generated image set $D_{S_{renewed}}$ and the annotations $B_{D_V}$, $D_{S_{renewed}}$ is the union of the generated fake thermal image set $D_{FT}$ and the inverted fake thermal image set $D_{FT_{Inv}}$, $D_V$ denotes the image set of color visible domain $V$, $B_{D_V}$ is the annotations of $D_V$. Noted that the renewed source domain $D_{S\_renewed}$ is with double number of $D_V$ and with annotations transferred from $D_V$.

Intuitively, we can train detector on annotated $D_{S\_renewed}$ directly, and apply it to target domain $T$. However, there still exists gap between $D_{S\_renewed}$ and $T$. Thus, we utilize an off-the-shelf Domain Adaptive Faster RCNN [7] (referred as DAF) to conduct a re-adaptation from $D_{S\_renewed}$ to $T$.

DAF [7] uses H-divergence to measure the divergence between data distribution of source domain and target domain. The authors formulate the object detection as a posterior learning problem in a probabilistic perspective, that is, $P(C, B|I)$, where I is the image, B is the bounding-box of an object and C is the category of the object. Based on the H-divergence measure and the probabilistic formulation, three adaptation component are proposed, i.e., image-level adaptation, instance-level adaptation and consistency regularization. Three adaptation components are trained jointly with adversarial learning.

## 3. EXPERIMENTS

In this section, various experiments are conducted to evaluate the effectiveness of the proposed method. In Section 3.1, we introduce the experiments setup including dataset, evaluation metric and implementation. In Section 3.2, we compare the proposed method with the state-of-art methods in accuracy performance. Finally, in Section 3.3, we analyze and discuss the impact of each module in ablation study.

### 3.1. Setup
*3.1.1. Dataset*
In order to evaluate the proposed method, we conduct experiments on Multi-spectral Object Detection dataset [9]. The Multi-spectral Object Detection dataset [9] is collected for autonomous vehicles. It consists of RGB, NIR, MIR, and FIR images and added ground truth labels. There are total 7,512 images (3,740 taken at daytime and 3,772 taken at nighttime). Bounding box coordinates and labels are consisted in the ground truth. Four different images are simultaneously captured and each object is annotated in the spectral images. In this dataset, five class objects (*bike*, *car*, *car_stop*, *color_cone*, *and person*) are labelled. In our experiments, the RGB images with annotations are set as source domain, and the FIR, i.e., thermal images, are set as target domain. The annotations of thermal images are not used during training process.

*3.1.2. Evaluation metric*

To assess the performance of object detector, we adopt widely used mean Average Precision (mAP) as evaluation criteria, which is calculated by recall and precision.
Recall (R) and Precision (P) are used to get AP value of each class. The mAP means that the mean value of AP for all categories. They are defined as follows:

$$AP = \int_0^1 P(R) dR \quad (5)$$

$$mAP = \frac{1}{N_{cls}} \sum_{i=1}^{N_{cls}} AP_i \quad (6)$$

where $N_{cls}$ represents the number of categories.

*3.1.3. Implementation details*
Our experiments are implemented on PyTorch [10] platform. For CycleGAN, the open source PyTorch version [11] is used. The CycleGAN is trained with 200 echoes. For the re-adaptation part, we use an open source PyTorch implementation [12]. Faster RCNN and DAF are both trained with 20 echoes and parameters are set as default.

### 3.2. Comparison with the state-of-art methods
In this section, we evaluate the detection performance quantitatively and qualitatively. In quantitative part, mAP of the Faster RCNN [2] trained on source data, the baseline also state-of-the-art method Domain Adaptive Faster RCNN [7] (referred as DAF) and our proposed method are compared. In qualitative part, we compare the proposed method to the state-of-the-art method DAF [7].
**Quantitative evaluation**.

Table 1 summarizes the experimental results of different methods. We compare the proposed method with Faster RCNN [2] trained on source data, and Domain Adaptive Faster RCNN [7] (referred as DAF). The DAF is trained on annotated source data and unlabelled target data. The proposed method is trained on generated images with annotations of original color visible domain. Faster RCNN trained on annotated target samples is taken as oracle. The proposed method achieved the mAP of 26.5%, while Faster RCNN (Non-adapted) achieved 1.4 %, and DAF achieved 19.4 %. Our method outperforms DAF with 8.8%.

**Table 1.** Comparison of the performance of compared methods. The bold numbers represent the best results

|  | Bike | Car | Car_stop | Color_cone | Person | mAP |
|---|---|---|---|---|---|---|
| Faster RCNN [2] | 1.5 | 0.7 | 0.5 | 0 | 4.1 | 1.4 |
| DAF [7] | 1.04 | 3.1 | 0.71 | 0.03 | 39.0 | 8.8 |
| Ours | **20.3** | **30.8** | **11.7** | **12.9** | **56.9** | **26.5** |
| Oracle | 67.9 | 81.3 | 52.6 | 64.5 | 76.6 | 68.6 |

**Qualitative Evaluation**.
From Figure 4, we can find that our proposed method detects more objects correctly than Faster RCNN and DAF.

### 3.3. Ablation study

In this subsection, we conduct an ablation study to analyze the effect of each proposed component of the whole pipeline on performance.

Table 2 provided the ablation performance of different configuration of each proposed component. Comparing configurations with CycleGAN based image translation to those with gray translation, it seems that configs with CycleGAN perform better. For example, config in the 7th row obtains mAP 12.6% while the 1st row obtains 1.4% and the 3rd row obtains 5.3%. Comparing configs with both image translation (gray or CycleGAN) and intensity inversion to configs with only image translation, those with intensity inversion yield obvious gain. For example, config in the 8th row obtains mAP 22.4% while the 7st row obtains 12.6%. Finally, configs with re-adaptation performs better than those without re-adaptation. For example, config in the 10th row obtains mAP 26.5% while the 8st row obtains 22.4%. From the above analysis, it is clearly that three proposed components, i.e., CycleGAN based image translation, intensity invertion and re-adaptation, are all necessary and yield performance gain.

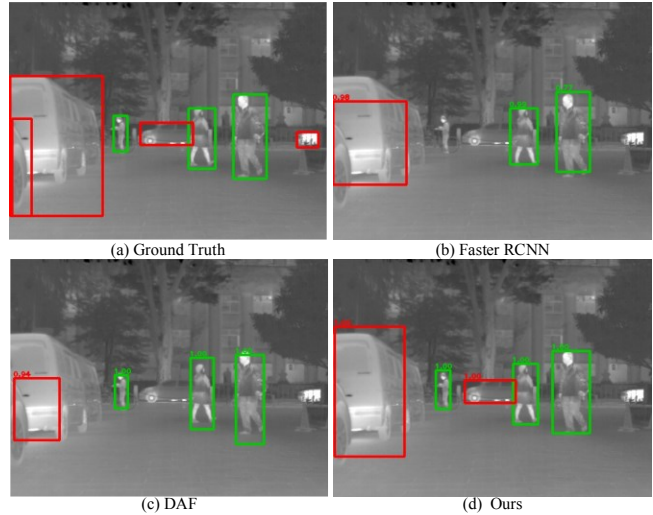

(a) Ground Truth (b) Faster RCNN
(c) DAF (d) Ours

**Figure 4**. Qualitative results of compared methods. The example image is 000726.png from FIR folder of Multi-spectral Object Detection dataset [9].

**Table 2.** Ablation study for each proposed component

| Image trans | Int-Inv[1] | R-A[2] | Bike | Car | Car_stop | Color_cone | person | mAP |
|---|---|---|---|---|---|---|---|---|
| No | - |  | 1.5 | 0.7 | 0.5 | 0 | 4.1 | 1.4 |
| No | - | √ | 1.04 | 3.1 | 0.71 | 0.03 | 39.0 | **8.8** |
| Gray |  |  | 0.94 | 4.1 | 1.2 | 2.03 | 18.3 | 5.3 |
| Gray | √ |  | 3.9 | 12.2 | 5.0 | 7.5 | 52.5 | 16.2 |
| Gray |  | √ | 1.5 | 2.1 | 2.5 | 7.2 | 50.7 | 12.8 |
| Gray | √ | √ | 3.0 | 7.3 | 4.7 | 17.0 | 61.2 | **18.6** |
| CycleGAN |  |  | 12.8 | 16.4 | 3.5 | 4.0 | 26.1 | 12.6 |
| CycleGAN | √ |  | 18.7 | 31.6 | 6.0 | 6.6 | 49.0 | 22.4 |
| CycleGAN |  | √ | 13.3 | 14.4 | 6.3 | 7.3 | 41.5 | 16.6 |
| CycleGAN | √ | √ | 20.3 | 30.8 | 11.7 | 12.9 | 56.9 | **26.5** |

Note: 1. Int-Inv indicates intensity inversion; 2. R-A means Re-adaptation

### 4. CONCLUSIONS

In this paper, we proposed an unsupervised image-generation enhanced adaptation method for object detection in thermal images. Two modules are included. The image-generation module is to generate simulated fake thermal images that are similar to the target images. And the re-adaptation module is to reduce the gap between generated intermediate domain and the thermal target domain. The presented experimental results demonstrate that the proposed method outperforms the state-of-the-art greatly. In the future, we will focus on methods for generating more similar thermal images from color visible images and compact end-to-end adaptive models.

### ACKNOWLEDGMENTS
This work is financially supported by National Natural Science Foundation of China (Nos. 61771471, 61401463,

## REFERENCES

[1] R. Gade and T. B. Moeslund, "Thermal cameras and applications: a survey," *Machine vision and applications,* vol. 25, pp. 245-262, 2014.

[2] S. Ren, K. He, R. Girshick, and J. Sun, "Faster R-CNN: Towards Real-Time Object Detection with Region Proposal Networks," *IEEE Transactions on Pattern Analysis and Machine Intelligence,* vol. 39, pp. 1137-1149, 2017.

[3] W. Liu, D. Anguelov, D. Erhan, C. Szegedy, S. Reed, C.-Y. Fu*, et al.*, "Ssd: Single shot multibox detector," in *European conference on computer vision*, 2016, pp. 21-37.

[4] J. Redmon, S. Divvala, R. Girshick, and A. Farhadi, "You only look once: Unified, real-time object detection," in *Proceedings of the IEEE conference on computer vision and pattern recognition*, 2016, pp. 779-788.

[5] C. Herrmann, M. Ruf, and J. Beyerer, "CNN-based thermal infrared person detection by domain adaptation," in *Autonomous Systems: Sensors, Vehicles, Security, and the Internet of Everything*, 2018, p. 1064308.

[6] T. Guo, C. P. Huynh, and M. Solh, "Domain-Adaptive Pedestrian Detection in Thermal Images," in *2019 IEEE International Conference on Image Processing (ICIP)*, 2019, pp. 1660-1664.

[7] Y. Chen, W. Li, C. Sakaridis, D. Dai, and L. Van Gool, "Domain Adaptive Faster R-CNN for Object Detection in the Wild," *computer vision and pattern recognition,* pp. 3339-3348, 2018.

[8] J.-Y. Zhu, T. Park, P. Isola, and A. A. Efros, "Unpaired image-to-image translation using cycle-consistent adversarial networks," in *Proceedings of the IEEE international conference on computer vision*, 2017, pp. 2223-2232.

[9] K. Takumi, K. Watanabe, Q. Ha, A. Tejero-De-Pablos, Y. Ushiku, and T. Harada, "Multispectral object detection for autonomous vehicles," in *Proceedings of the on Thematic Workshops of ACM Multimedia 2017*, 2017, pp. 35-43.

[10] A. Paszke, S. Gross, F. Massa, A. Lerer, J. Bradbury, G. Chanan*, et al.*, "PyTorch: An imperative style, high-performance deep learning library," in *Advances in Neural Information Processing Systems*, 2019, pp. 8024-8035.

[11] (2018, 2020/1/3). *Image-to-Image Translation in PyTorch*. Available: https://github.com/junyanz/pytorch-CycleGAN-and-pix2pix

[12] (2017, 2020/1/3). *An unofficial implementation of 'Domain Adaptive Faster R-CNN for Object Detection in the Wild '*. Available: https://github.com/tiancity-NJU/da-faster-rcnn-PyTorch